CAMBRIDGE UNIVERSITY PRESS

**METHODS PAPER**

# Data-driven integration kernels for interpretable nonlocal operator learning

Savannah L. Ferretti[1], Jerry Lin[2], Sara Shamekh[3], Jane W. Baldwin[1,4], Michael S. Pritchard[5] and Tom Beucler[6,7]

[1]Department of Earth System Science, University of California Irvine, Irvine, USA
[2]Department of Computing and Data Science, Boston University, Boston, USA
[3]The Center for Atmosphere Ocean Science, New York University, New York, USA
[4]Lamont-Doherty Earth Observatory, Palisades, USA
[5]NVIDIA Corporation, Santa Clara, USA
[6]Faculty of Geosciences and Environment, University of Lausanne, Lausanne, Switzerland
[7]Expertise Center for Climate Extremes, University of Lausanne, Lausanne, Switzerland
**Corresponding author:** Savannah L. Ferretti; Email: savannah.ferretti@uci.edu.

## Abstract

Machine learning models can represent climate processes that are nonlocal in horizontal space, height, and time, often by combining information across these dimensions in highly nonlinear ways. While this can improve predictive skill, it makes learned relationships difficult to interpret and prone to overfitting as the extent of nonlocal information grows. We address this challenge by introducing data-driven integration kernels, a framework that adds structure to nonlocal operator learning by explicitly separating nonlocal information aggregation from local nonlinear prediction. Each spatiotemporal predictor field is first integrated using learnable kernels (defined as continuous weighting functions over horizontal space, height, and/or time), after which a local nonlinear mapping is applied only to the resulting kernel-integrated features and optional local inputs. This design confines nonlinear interactions to a small set of integrated features and makes each kernel directly interpretable as a weighting pattern that reveals which horizontal locations, vertical levels, and/or past timesteps contribute most to the prediction. We demonstrate the framework for South Asian monsoon precipitation using a hierarchy of neural network models with increasing structure, including baseline, nonparametric kernel, and parametric kernel models. Across this hierarchy, kernel models achieve near-baseline performance with far fewer trainable parameters, indicating that much of the relevant nonlocal information can be captured through a small set of interpretable integrations when appropriate structural constraints are imposed.

## Impact Statement

Many geophysical processes depend on nonlocal structure in horizontal space, height, and time, yet most data-driven models encode these dependencies implicitly, limiting interpretability and robustness. This work introduces data-driven integration kernels, a framework that makes nonlocal influence explicit by separating how information is aggregated from how local predictions are made. Because kernels define physically interpretable weighting patterns, they allow nonlocal structure to be analyzed and compared alongside predictive

This research article was awarded Open Data and Open Materials badges for transparent practices. See the Data Availability Statement for details.

performance, revealing whether different modeling choices rely on similar forms of nonlocal dependence. The resulting kernel-integrated features provide a direct pathway for informing and constraining physically interpretable parameterizations derived from data-driven models.

## 1. Introduction

Geophysical processes are inherently nonlocal, with local outcomes depending on conditions across neighboring horizontal locations, the vertical column, and past timesteps. A growing body of machine learning research addresses this structure by learning operators (Kovachki et al., 2024) that map entire, high-dimensional input fields to local outputs. Such models have been used to learn nonlocal processes in weather and climate, and have achieved strong predictive performance across a range of applications (Pathak et al., 2022; Bonev et al., 2025; Watt-Meyer et al., 2025). These operator-learning approaches can flexibly combine information across horizontal space, height, and time, but nonlocal contributions are typically encoded implicitly within large parameter sets, making it difficult to identify which spatial scales, vertical levels, or memory timescales are most influential. As a result, adding broader nonlocal context can increase model complexity without providing clearer insight into *how* nonlocal information is actually used.

A range of strategies has been proposed to manage this complexity. Dimensionality-reduction approaches such as principal-component analysis and encoder–decoder architectures (including autoencoders) compress high-dimensional fields into lower-dimensional latent representations, but these representations often reflect architectural or statistical choices rather than physically meaningful structure (Monahan et al., 2009; King et al., 2025; Wani, 2025). Post-hoc explainability methods aim to attribute predictions to influential regions or features, but depend on both the trained model and the attribution technique. As a result, they can mix contributions across predictors and dimensions and yield sample-dependent attributions that are difficult to interpret physically (Silva and Keller, 2024; O'Loughlin et al., 2025). Because these approaches rely on additional models fitted after training, they can compound uncertainty, reinforce prior expectations, and fail to provide stable summaries of nonlocal influence.

We introduce *integration kernel learning*, which addresses these requirements by representing a model's nonlocal behavior through learned kernels that weight information across horizontal space, height, and/or time prior to prediction. By explicitly separating nonlocal integration from local nonlinear mapping, the framework constrains the operator class while preserving flexibility in local response. The learned kernels yield continuous, coordinate-aware summaries of nonlocal influence, embedding interpretability directly into the operator without sacrificing predictive skill.

**Key contributions of this work include:**

- We introduce integration kernel learning as an interpretable framework for representing nonlocal operators using continuous weighting functions over horizontal space, height, and time.
- We show that separating nonlocal integration from local nonlinear mapping regularizes the operator class and yields interpretable kernels.
- We develop a hierarchy of models, from unconstrained neural networks to nonparametric and parametric kernel models, to quantify trade-offs between skill, complexity, and interpretability.
- We apply the framework to South Asian monsoon precipitation as a case study, showing that kernel models retain most of the predictive skill of full-field models while revealing key dependencies.

## 2. Methodology

We represent nonlocal dependencies in continuous geophysical data using a two-step procedure. First, predictor fields are integrated over horizontal space, height, and/or time using learnable kernels to form lower-dimensional features. Second, these kernel-integrated features are passed to a local nonlinear mapping to predict the target variable. We use a spatiotemporal coordinate system standard in atmospheric applications, but the formulation is readily generalized to other nonlocal processes and coordinate systems.

### 2.1. Nonlocal operators via integration kernels

Let $\mathbf{x}$ denote horizontal location, $p$ denote pressure, and $t$ denote time. We aim to predict a target variable (e.g., precipitation) $y(\mathbf{x}_0, t_0)$ at location $\mathbf{x}_0$ and time $t_0$ using predictor fields (e.g., temperature and humidity) $\{\varphi_i(\mathbf{x}, p, t)\}_{i=1}^{N_\varphi}$, where $N_\varphi$ is the number of predictor variables. Predictor fields are defined over horizontal space, pressure, and time, while the target variable is defined only over horizontal space and time.

We view the mapping from predictor fields to the target as a nonlocal operator $\mathcal{L}$ and approximate:

$$y(\mathbf{x}_0, t_0) \approx \mathcal{L}[\{\varphi_i\}](\mathbf{x}_0, t_0), \tag{2.1}$$

where $\{\varphi_i\}$ denotes the collection of all predictor fields. A purely local model would restrict $y(\mathbf{x}_0, t_0)$ to depend only on $\varphi_i(\mathbf{x}_0, p_0, t_0)$, where $p_0$ denotes the vertical level nearest the surface, which is insufficient for quantities such as precipitation that depend on nonlocal structure (Emanuel et al., 1994). We therefore allow $\mathcal{L}$ to depend on a surrounding neighborhood in horizontal space, height, and time.

To parameterize this neighborhood dependence, we use *integration kernels*. For each predictor field $\varphi_i$, integration kernels define linear operators that aggregate the field over specified domains in horizontal space, height, and/or time relative to the prediction point. Each operator produces a kernel-integrated feature, and multiple kernels for the same predictor may differ in their domains or weighting patterns, yielding distinct aggregated summaries of $\varphi_i$. The collection of these features provides a low-dimensional representation of the nonlocal information used in the subsequent prediction step.

Formally, the $\ell^{\text{th}}$ kernel-integrated feature associated with predictor $\varphi_i$ is defined as:

$$\widehat{\varphi}_i^{(\ell)}(\mathbf{x}_0, t_0) = \int_{t_0-\tau_{\max}}^{t_0} \int_0^{p_s} \int_{\mathcal{D}} k_i^{(\ell)}(\mathbf{x}, p, t; \mathbf{x}_0, t_0)\, \varphi_i(\mathbf{x}, p, t)\, \mathrm{d}\mathbf{x}\, \mathrm{d}p\, \mathrm{d}t, \tag{2.2}$$

where $\mathcal{D}$ denotes a neighborhood of horizontal locations around $\mathbf{x}_0$, $p \in [0, p_s]$ is the pressure coordinate with $p = 0$ at the top of the atmosphere and $p = p_s$ at the surface, and $\tau_{\max} > 0$ sets the temporal memory. The temporal integral is restricted to $t \le t_0$ to prevent leakage of future information. The kernel $k_i^{(\ell)}(\mathbf{x}, p, t; \mathbf{x}_0, t_0)$ assigns weights to $\varphi_i$ at $(\mathbf{x}, p, t)$ relative to the prediction point $(\mathbf{x}_0, t_0)$.

### 2.2. Approximating the nonlocal operator with kernel-integrated features

Equation 2.2 defines kernel-integrated features $\widehat{\varphi}_i^{(\ell)}(\mathbf{x}_0, t_0)$ that summarize nonlocal information in each predictor field $\varphi_i$. In addition to these features, we allow optional local inputs $\boldsymbol{\psi}(\mathbf{x}_0, t_0)$ (e.g., surface fluxes or land properties) evaluated directly at the prediction point. We can model the target variable as a function of both:

$$y(\mathbf{x}_0, t_0) \approx \underbrace{F\left(\{\widehat{\varphi}_i^{(\ell)}(\mathbf{x}_0, t_0)\}_{i,\ell}, \boldsymbol{\psi}(\mathbf{x}_0, t_0)\right)}_{\text{local in } \hat{\varphi}} \approx \underbrace{F\left(\{\mathcal{K}_i^{(\ell)}[\varphi_i](\mathbf{x}_0, t_0)\}_{i,\ell}, \boldsymbol{\psi}(\mathbf{x}_0, t_0)\right)}_{(\text{local operator } F)\,\circ\,(\text{nonlocal integrations } \mathcal{K}_i^{(\ell)})}. \tag{2.3}$$

Here, $\mathcal{K}_i^{(\ell)}$ denotes the nonlocal integration operator defined by the kernel $k_i^{(\ell)}$ in Equation 2.2. Each operator $\mathcal{K}_i^{(\ell)}$ acts on a single predictor field $\varphi_i$, while nonlinear interactions among predictors are learned by $F$, which we implement using neural networks. The original operator acting on full spatiotemporal fields $(\mathbf{x}, p, t)$ is thus approximated by the composition $F \circ \mathcal{K}$, with $F$ operating only on the kernel-integrated feature space at $(\mathbf{x}_0, t_0)$.

This factorization confines nonlocal aggregation to the kernel stage and restricts interactions across predictors to a finite set of kernel-integrated features, reducing the effective dimensionality of the mapping and enabling direct interpretation of the learned kernels as weighting functions over horizontal space, height, and/or time.

### 2.3. Discrete kernel operators and learning from data

To learn the integration kernels from discrete data, we sample each predictor field $\varphi_i(\mathbf{x},p,t)$ on a horizontal grid $\{\mathbf{x}_n\}_{n=1}^{N_\mathbf{x}} \subset \mathcal{D}$, a pressure grid $\{p_m\}_{m=1}^{N_p} \subset [0,p_s]$, and past timesteps $\{t_r\}_{r=1}^{N_t} \subset [t_0 - \tau_{\max}, t_0]$. Quadrature weights $\Delta A_{n,m,r}$, $\Delta p_{n,m,r}$, and $\Delta t_{n,m,r}$ represent horizontal area, pressure, and temporal integration elements and allow for irregular grids or custom integration schemes, though they do not, in general, guarantee mesh-invariant representations. With this discretization, the kernel-integrated feature in Equation 2.2 is approximated by the weighted sum:

$$\widehat{\varphi}_i^{(\ell)}(\mathbf{x}_0,t_0) \approx \sum_{r=1}^{N_t}\sum_{m=1}^{N_p}\sum_{n=1}^{N_\mathbf{x}} k_{i,n,m,r}^{(\ell)}(\mathbf{x}_0,t_0)\, \varphi_i(\mathbf{x}_n,p_m,t_r)\, \Delta A_{n,m,r}\, \Delta p_{n,m,r}\, \Delta t_{n,m,r}, \tag{2.4}$$

where $k_{i,n,m,r}^{(\ell)}$ are the discretized kernel weights.

For each predictor field $\varphi_i$, the kernel weights $\{k_{i,n,m,r}^{(\ell)}\}$ form a tensor $\mathbf{K}_i \in \mathbb{R}^{L_i \times N_\mathbf{x} \times N_p \times N_t}$, with one kernel per feature index $\ell$. Each kernel is normalized after discretization so that it integrates to one over the chosen horizontal space, height, and time domain, and thus acts as a weighting pattern when forming features. Kernel weights are not required to be positive, allowing kernels to emphasize, suppress, or oppose contributions from different regions.

In practice, values of $\varphi_i$ that are undefined at a given location (i.e., at pressure levels exceeding the local surface pressure) are excluded from the summation in Equation 2.4 via a binary validity mask. Masked values are set to zero in standardized space (corresponding to the training set mean) and omitted from the quadrature sum, ensuring they do not bias kernel-integrated features. In such cases, the effective kernel integral over the valid vertical domain may be less than one, particularly at high-elevation locations where surface pressure is lower. Rather than renormalizing kernels locally, we provide the validity mask to the downstream mapping, allowing it to learn and account for variations in effective column depth during training.

Depending on the application, kernels may be defined over all three dimensions or restricted to a subset, corresponding to limiting the integration domain in Equation 2.4 and constraining the structure of $\mathbf{K}_i$. After computing all kernel-integrated features, they are concatenated with any local inputs $\boldsymbol{\psi}(\mathbf{x}_0,t_0)$ to form the feature vector used for prediction. Once the integration domains are specified, the most general kernel representation is a nonparametric one, with all weights $k_{i,n,m,r}^{(\ell)}$ learned directly. Because kernels are shared across all samples, they encode a consistent description of how each predictor field contributes nonlocally to the prediction. However, strong correlations in the data can render such kernels underdetermined and prone to overfitting.

We therefore also consider parametric kernel families that restrict the integration stage to simple functional shapes. These include kernels that emphasize a preferred coordinate value (Gaussian), limited multimodal structure (mixture-of-Gaussians), uniform averaging over an interval (top-hat), and monotonic decay away from a boundary or reference point (exponential). By constraining the form of the nonlocal aggregation, parametric kernels reduce the number of learnable parameters even further while yielding more interpretable summaries of nonlocal influence. In practice, these kernel families may be applied uniformly across predictors or combined in predictor-specific mixtures. The explicit functional forms used in this study are provided in Appendix A.

## 3. Example implementation

### 3.1. Motivation

Parameterizing South Asian monsoon rainfall as a function of thermodynamic input variables provides a stringent, practically relevant testbed for methods that learn nonlocal relationships in atmospheric data. Monsoon precipitation has strong societal and hydrological impacts (Gadgil and Gadgil, 2006), yet global climate models continue to struggle with its intensity, spatial organization, and variability (Douville et al., 2021), in part due to uncertainty in how to represent the precipitation formation process as a function of local and nonlocal thermodynamic structure. These challenges make the monsoon a reasonably representative setting for assessing whether integration kernels can identify how atmospheric structure across pressure levels, spatial neighborhoods, and past timesteps contributes to precipitation.

### 3.2. Implementation framework

We apply integration kernel learning to predict precipitation over the South Asian monsoon region (5–25°N, 60–90°E) during June–August of 2000–2020. Predictions are made using multiple thermodynamic predictor fields together with a small set of local inputs.

The thermodynamic predictor fields are relative humidity (RH), equivalent potential temperature ($\theta_e$), and saturated equivalent potential temperature ($\theta_e^*$), which are relevant for South Asian monsoon rainfall variability (Ferretti et al., 2025). Local inputs (evaluated only at the prediction point) are sensible heat flux, latent heat flux, and land fraction. ERA5 reanalysis (Hersbach et al., 2023a, 2023b) provides physically consistent, well-resolved vertical profiles over the South Asian monsoon region, particularly for lower-tropospheric temperature and moisture (Hersbach et al., 2020; Johnston et al., 2021). Accordingly, all predictor fields and local inputs are extracted from hourly ERA5 data on its native 0.25° × 0.25° grid. Thermodynamic predictors are constructed from surface pressure and from temperature and specific humidity on pressure levels between 1,000 and 500 hPa. Precipitation targets are taken from half-hourly IMERG V06 (Huffman et al., 2019) on its native 0.1° × 0.1° grid, and are aggregated to hourly resolution using centered averages. Prior to model training, both datasets are conservatively regridded to a common 1.0° × 1.0° grid, consistent with the horizontal scales at which precipitation is typically parameterized in large-scale atmospheric models.

To evaluate the role of nonlocal information, we prescribe in advance the extent of spatial, vertical, and temporal context available to the model when predicting precipitation at a location $(\mathbf{x}_0, t_0)$. A model is considered *nonlocal* along a given dimension if it is allowed to depend on predictor values within a fixed domain around $(\mathbf{x}_0, t_0)$ along that dimension. Horizontal nonlocality corresponds to a 3 × 3 neighborhood of adjacent grid cells centered on $\mathbf{x}_0$ (9 grid cells total), vertical nonlocality corresponds to pressure levels between 1,000 and 500 hPa (16 levels total), and temporal nonlocality corresponds to access to the current timestep together with the six preceding timesteps (7 hours total). Models that are local along a given dimension depend only on predictor values within the current grid cell, at the current timestep, and at the lowest pressure level above the local surface. All models are trained and evaluated on an identical set of horizontal locations and timesteps for which the full required nonlocal context is available. Prediction locations lacking the necessary horizontal, vertical, or temporal context, due to proximity to domain boundaries or insufficient temporal coverage, are excluded from all experiments.

We use a chronological split with data from 2000–2014 for training, 2015–2017 for validation, and 2018–2020 for testing. All variables except land fraction (which is time-invariant and bounded in $[0,1]$) are standardized using training set means and standard deviations. Precipitation is transformed using $\log(1+y)$ prior to standardization to reduce the influence of extremes and improve training stability, and all losses and reported skill metrics are evaluated in this standardized log-transformed precipitation space.

Figure 1 provides a schematic overview of the kernel modeling framework implemented in this study.

### 3.3. Model hierarchy

To isolate the role of nonlocal structure and to assess the value of integration kernels, we construct a three-level model hierarchy in which the representation of nonlocal information is varied while all other components are held fixed. Across all models, the same downstream neural network is used (see Appendix B) to map inputs to precipitation, ensuring that differences in performance arise solely from how nonlocal information is encoded.

At the first level, we train *baseline* models that ingest predictor-field values from the prescribed horizontal space, height, and time domain. For each predictor $\varphi_i$, values within this fixed domain are flattened and concatenated with the local inputs before being passed to the neural network. In these models, nonlocal aggregation and nonlinear interactions are learned jointly within the network weights, providing maximal flexibility but limited interpretability.

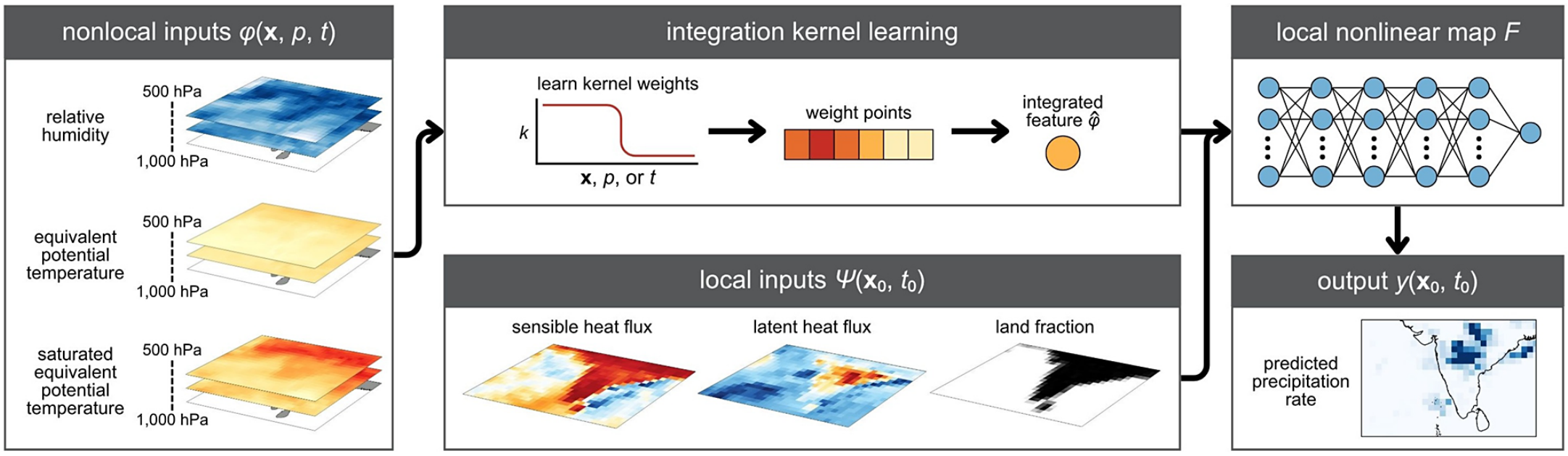

***Figure 1.*** *Schematic of integration kernel learning. Learned kernels summarize predictor fields across horizontal space, height, and/or time into features, which are combined with local inputs and passed to a downstream nonlinear model to predict the local output.*

At the second level, we train *nonparametric kernel* models that explicitly separate nonlocal aggregation from local nonlinear prediction. Each predictor field is summarized by a set of learned kernel-integrated features, which are concatenated with the local inputs and passed to the same neural network used in the baseline models. This structure reduces the effective number of trainable parameters and confines nonlinear interactions to a reduced feature space.

At the third level, we train *parametric kernel* models that further constrain the aggregation stage by restricting kernels to simple functional families (e.g., Gaussian, mixture-of-Gaussians, top-hat, or exponential; see Appendix A). As in the nonparametric case, predictor fields are integrated using these kernels and then mapped to precipitation through the same downstream network. We consider both configurations that apply a single kernel family across all predictors and those that allow predictor-specific choices. In this sense, the selected parametric families are informed by the qualitative structure of the nonparametric kernels for this application, but the framework itself does not impose these choices, and alternative functional forms can be adopted to better match different datasets or physical regimes.

## 4. Results

We evaluate the integration kernel approach by (1) examining how predictive performance depends on the representation of nonlocal structure and (2) analyzing the kernel structure recovered by the models. All results are reported on the test set at the common prediction points defined in Section 4.2.

Figure 2 summarizes predictive performance across the model hierarchy introduced in Section 4.3, using test set $R^2$ and mean squared error (MSE), both computed in standardized log-transformed precipitation space. $R^2$ provides a scale-independent measure of explained variance, while MSE quantifies absolute error aligned with the training objective. Table C1 reports corresponding uncertainties from bootstrap resampling and variability across random seeds, which are uniformly small ($O(10^{-3})$), indicating robust performance differences across sampling and initialization.

Considering only the baseline models in Figure 2, the fully local $(\mathbf{x}_0, p_0, t_0)$ and fully nonlocal $(\mathbf{x}, p, t)$ models define lower and upper bounds on skill for the prescribed nonlocal domain ($R^2 = 0.411$, MSE = 0.680, and $R^2 = 0.582$, MSE = 0.482, respectively). Intermediate models that introduce nonlocality along a single dimension provide an ablation of horizontal, vertical, and temporal contributions. Among these, the vertically nonlocal model ($R^2 = 0.528$, MSE = 0.546) lies closest to the upper bound, whereas the horizontally ($R^2 = 0.440$, MSE = 0.647) and temporally ($R^2 = 0.432$, MSE = 0.656) nonlocal models remain closer to the lower bound. This ordering is consistent with the dominant role of vertical thermodynamic structure in controlling convective precipitation, with horizontal and temporal context playing a secondary role at the scales considered here (Arakawa and Schubert, 1974). That this hierarchy within the baseline models reflects expected physical behavior provides reassurance that the learned representations capture physically meaningful sources of predictive skill, motivating our focus on developing kernels in the vertical dimension.

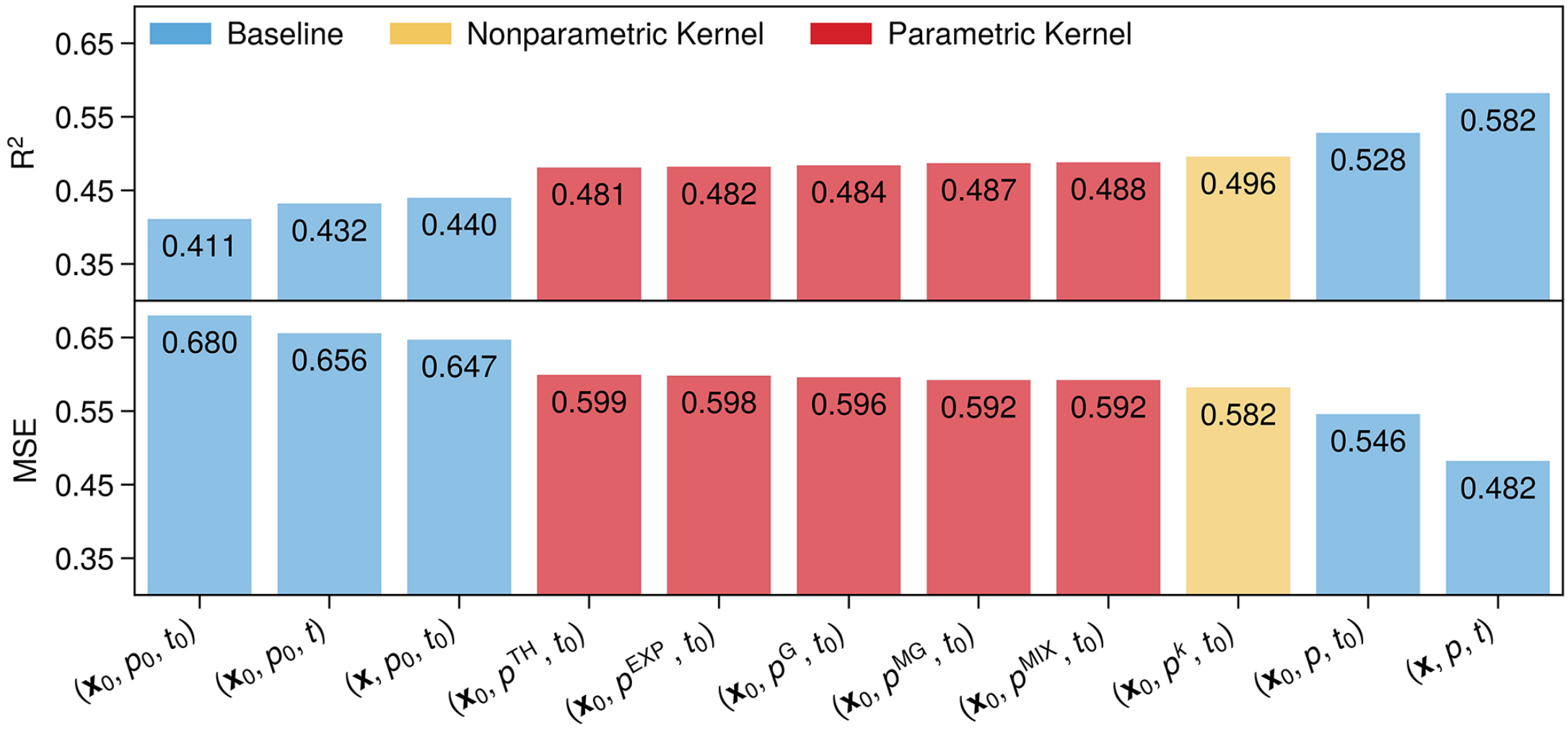

***Figure 2.*** *Test set $R^2$ (top) and MSE (bottom) for baseline (blue), nonparametric kernel (yellow), and parametric kernel (red) models, computed in standardized log-transformed precipitation space. Model labels indicate nonlocal dimensions, with subscript 0 denoting locality. Superscript k denotes nonparametric kernels, while TH, EXP, G, and MG denote top-hat, exponential, Gaussian, and mixture-of-Gaussians kernels (see Appendix A). MIX uses MG kernels for RH and $\theta_e^*$ and EXP for $\theta_e$.*

Relative to the vertically nonlocal baseline model, kernel models in Figure 2 show that the nonparametric kernel model achieves comparable performance despite operating on substantially reduced input representations ($R^2$ = 0.496 versus 0.528, MSE = 0.582 versus 0.546). The parametric kernel models perform slightly worse than the nonparametric kernel model ($R^2 \approx$ 0.481–0.488, MSE ≈ 0.592–0.599), with only minor spread across formulations, indicating limited sensitivity to the specific parametric form once vertical nonlocality is enforced. The remaining degradation of the parametric kernel models relative to the nonparametric kernel model reflects stronger functional constraints in the integration stage. Spatial composites for dry, typical, and wet conditions show consistent qualitative behavior across the vertically nonlocal baseline model, the best nonparametric kernel model, and the best parametric kernel model (Figure C1), demonstrating that these models reproduce ground truth patterns closely for various precipitation regimes.

Figure 3 shows that the learned vertical kernels exhibit distinct, predictor-specific structure across the lower troposphere, consistent with known physical controls on South Asian monsoon convection (Parker et al., 2016; Menon et al., 2018; Ahmed and Neelin, 2021; Gayatri et al., 2024; Ferretti et al., 2025). For RH, the nonparametric kernels emphasize both near-surface levels (approximately 900–1,000 hPa) and the lower free troposphere (roughly 650–500 hPa), reflecting the combined roles of boundary layer moisture supply and free-tropospheric humidity in regulating convective intensity and precipitation efficiency. The $\theta_e$ kernels exhibit broadly positive weighting through the lower troposphere, with a localized negative contribution near 600 hPa, indicating sensitivity to the contrast between boundary layer parcel energy and lower free-tropospheric conditions, rather than to $\theta_e$ at any single level. The $\theta_e^*$ kernels show alternating positive and negative contributions in the lower free troposphere, indicating sensitivity to the vertical distribution of stability and entrainment-driven dilution that determines whether boundary layer buoyancy can be maintained during ascent (Ahmed and Neelin, 2018).

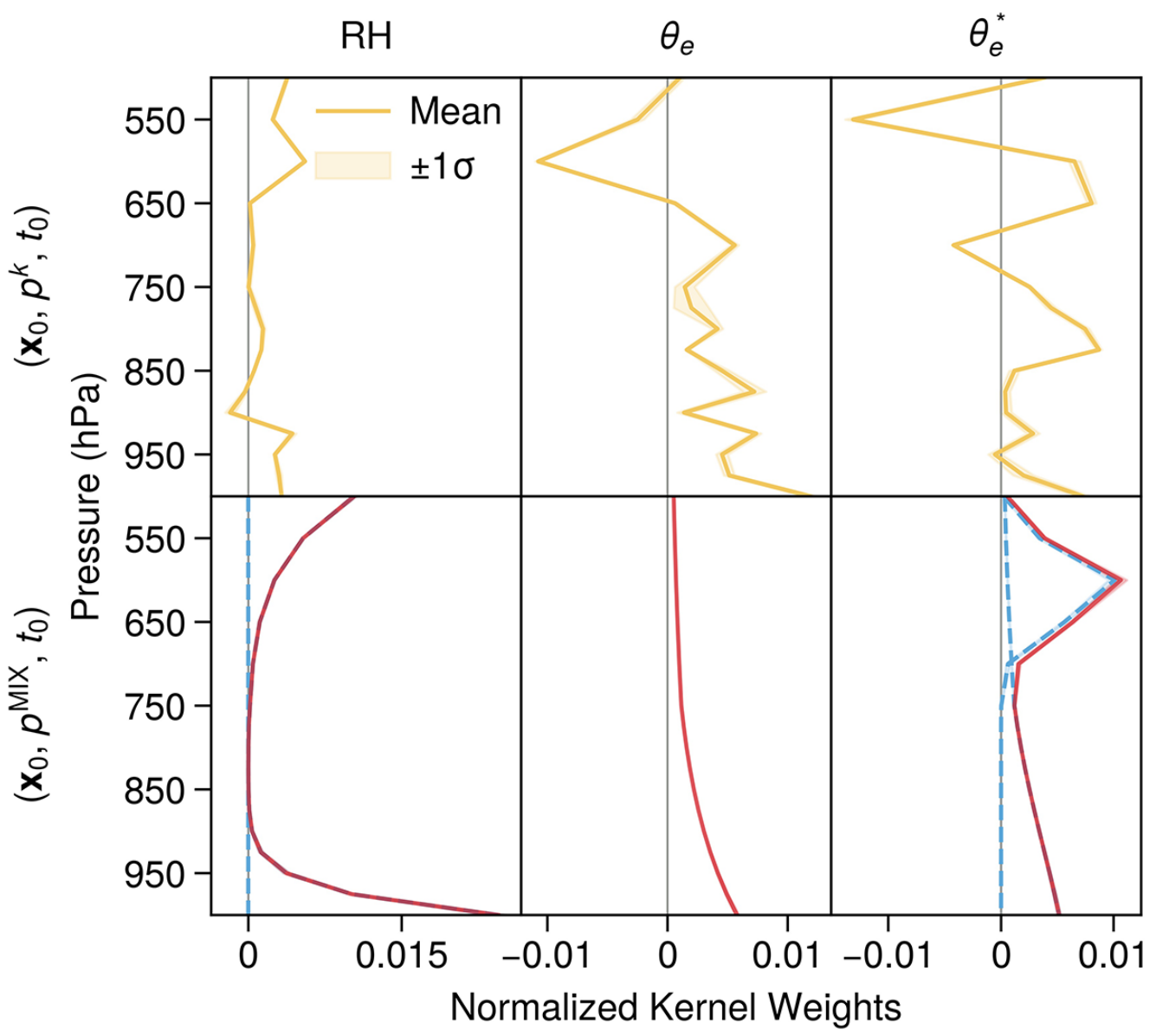

***Figure 3.*** *Learned vertical kernels for predictor fields from the two best kernel models in Figure 2. Solid lines show mean weights across seeds (n = 3), shaded by standard deviation. Absolute magnitude indicates pressure level importance, with sign indicating influence direction. For mixture-of-Gaussians kernels, dashed blue lines show the two Gaussian components and the solid red line their sum. The $\theta_e$ kernel in the bottom row uses the exponential kernel and therefore has no component decomposition.*

Parametric kernel models retain these dominant vertical sensitivities while imposing stronger structural constraints. In our best parametric model, RH and $\theta_e^*$ are each represented using mixture-of-Gaussians kernels (sums of two Gaussians), while $\theta_e$ is represented using an exponential kernel, choices motivated by the qualitative structure of the nonparametric kernels described above. Together, these kernel forms capture the primary boundary layer and lower free-tropospheric sensitivities while suppressing fine-scale variability, with higher-order oscillations and sign-changing structure absorbed into the downstream nonlinear mapping. The parametric $\theta_e^*$ kernel in particular collapses onto a dominant positive peak near 600 hPa, where the moisture deficit $(\theta_e^* - \theta_e)$ becomes increasingly important for regulating entrainment-driven dilution of rising parcels, consistent with the positive RH weighting at similar levels. That the parametric and nonparametric $\theta_e^*$ kernels differ most in shape, which reflects both the structural constraints of the mixture-of-Gaussians kernel and underdetermination, since multiple kernel profiles can achieve similar predictive performance. Kernels for all predictors and kernel models are shown in Figure C2.

## 5. Conclusion

We introduced data-driven integration kernels as an interpretable framework for learning nonlocal operators by separating nonlocal linear aggregation from local nonlinear prediction. This structure constrains model complexity, reduces dimensionality, and yields kernels that are directly interpretable as weighting patterns over horizontal space, height, and/or time. When applied to South Asian monsoon precipitation, we find that most of the predictive skill gained from nonlocal information arises from vertical structure. This is consistent with established understanding that vertical thermodynamic profiles exert primary control on the generation of deep convective rainfall at the scales considered here, and it provides reassurance that the framework captures physically meaningful sources of predictive skill. At the same time, a key methodological result is that kernel models that restrict aggregation to this dimension retain performance close to the strongest baselines while using substantially fewer input features, with parametric kernels providing additional regularization at some cost to predictive skill. Beyond prediction, kernel-integrated features provide compact summaries of nonlocal influence that are well suited for symbolic regression, offering a potential route toward simpler analytic approximations of the learned operators. This may enable future work to more directly compare and interpret nonlocal mechanisms and assess their variation across weather regimes.

**Acknowledgements.** We thank Fiaz Ahmed and Eric Wengrowski for their contributions during the early stages of this work, Jared Sexton for his helpful comments on the manuscript prior to submission, and Jo Lécuyer for developing an early version of the data-driven integration kernels during a University of Lausanne Master's internship.

**Author contribution.** T.B. developed the initial research concept and provided project supervision. S.L.F. conducted the investigation, including data curation, software development, formal analysis, validation, and visualization. Methodology was developed by T.B. with technical support from S.L.F. and J.L. Funding acquisition and project administration were led by T.B. and M.S.P., with additional administrative support from J.W.B. and resources provided by M.S.P. The manuscript was drafted by S.L.F. and T.B. and revised by J.L., S.S., J.W.B., and M.S.P. All authors approved the final version.

**Competing interests.** The authors declare no competing interests.

**Data availability statement.** ERA5 (Hersbach et al., 2023a, 2023b) and IMERG V06 (Huffman et al., 2019) data used in this study are publicly available in analysis-ready, cloud-optimized formats via the LEAP Data Catalog (https://catalog.leap.columbia.edu/feedstock/arco-era5) and the Microsoft Planetary Computer Data Catalog (https://planetarycomputer.microsoft.com/dataset/gpm-imerg-hhr), respectively. All code used for data retrieval and processing, model training and inference, and visualization is archived on Zenodo (https://doi.org/10.5281/zenodo.20562314).

**Ethical statement.** The research meets all ethical guidelines, including adherence to the legal requirements of the study country.

**Funding statement.** S.L.F. was supported by the National Science Foundation (NSF) Science and Technology Center (STC) Learning the Earth with Artificial Intelligence and Physics (LEAP) (2019625-STC) and the U.S. Department of Energy (DOE) Advanced Scientific Computing Research (ASCR) Program (DE-SC0022255). J.L. was supported by the DOE Office of Science through the Program for Climate Model Diagnosis and Intercomparison (PCMDI). S.S. was supported by the DOE Office of Biological and Environmental Research (BER) (DE-SC0025456). M.S.P. acknowledges support from NVIDIA Corporation. J.W.B. was supported by the National Aeronautics and Space Administration (NASA) New Investigator Program in Earth Science (NIP) (80NSSC21K1735). T.B. was supported by the Swiss State Secretariat for Education, Research, and Innovation (SERI) under the Horizon Europe AI4PEX Project (Grant Agreement ID 101137682; SERI No. 23.00546). Computational resources were provided by Perlmutter at the National Energy Research Scientific Computing Center (NERSC), a DOE User Facility (Project m4334).

## Appendix A: Parametric kernel families

The parametric kernel families used to construct the nonlocal integration operators in Section 3.3 are defined below. As with nonparametric kernels, they are evaluated on discrete grids along one or more integration coordinates and then normalized using the quadrature-based procedure described therein.

All integration coordinates are internally rescaled to [–1, 1] for numerical stability and consistent parameterization across dimensions. For horizontal neighborhoods centered at the prediction location, –1 and +1 correspond to the edges of the local spatial domain. In the vertical dimension, –1 corresponds to the top of the column (500 hPa) and +1 to the bottom (1,000 hPa), while in time $-1$ corresponds to the earliest point in the integration window and $+1$ to the prediction time. Kernel parameters are learned in this normalized coordinate space and mapped back to physical coordinates for interpretation.

Each integration kernel is defined as a function of a generic coordinate $s \in \Omega_s$, where $s$ represents horizontal location ($s = \mathbf{x}$), pressure ($s = p$), or time ($s = t$). The domain $\Omega_s$ is one-dimensional for $s \in \{p, t\}$ and two-dimensional for $s = \mathbf{x}$, with discretization inherited from the underlying data grid. When kernels act along multiple coordinates, separable kernels are formed as products of the corresponding one- or two-dimensional components.

**Gaussian Kernel**. A Gaussian kernel localizes integration around a preferred coordinate value:

$$k^{G}(s) = \exp\left(-\frac{\|s-\mu\|^2}{2\sigma^2}\right), \tag{A.1}$$

where $\mu$ is the kernel center, $\sigma > 0$ controls the width, and $\|\cdot\|$ denotes the Euclidean norm when $s = \mathbf{x}$.

**Mixture-of-Gaussians Kernel.** To allow limited bimodal or antagonistic structure, we represent mixture-of-Gaussians kernels using two Gaussian components with learned scalar weights $w_1, w_2$ that may be positive or negative. Negative weights enable one component to suppress the influence of the other. The resulting kernel is defined as:

$$k^{MG}(s) = w_1 \exp\left(-\frac{\|s-\mu_1\|^2}{2\sigma_1^2}\right) + w_2 \exp\left(-\frac{\|s-\mu_2\|^2}{2\sigma_2^2}\right), \tag{A.2}$$

where $\mu_1, \mu_2$ are component centers, $\sigma_1, \sigma_2 \in [0.1, 2.0]$ are component widths (clamped for numerical stability), and $\|\cdot\|$ denotes the Euclidean norm when $s = \mathbf{x}$.

**Top-Hat Kernel.** The top-hat kernel represents uniform averaging over a bounded interval of the integration domain. It is implemented as a product of two sigmoid functions for differentiability:

$$k^{TH}(s) = \left(1 + \exp\left(-\frac{s-a}{\varepsilon}\right)\right)^{-1}\left(1 + \exp\left(-\frac{b-s}{\varepsilon}\right)\right)^{-1}, \tag{A.3}$$

where $\varepsilon = 0.02$ is a fixed sharpness parameter, $a = \min(l, u)$ and $b = a + \widetilde{w}$ are the effective lower and upper bounds, and $l, u$ are learned parameters. The effective width $\widetilde{w}$ is derived from the raw learned width $w = \max(l, u) - \min(l, u)$ via soft clamping: $\widetilde{w} = w$ for $w \leq 1.5$, and $\widetilde{w} = 1.5\tanh(w/1.5)$ otherwise, preventing the kernel from spanning the full normalized domain.

**Exponential Kernel.** The exponential kernel represents influence that decays away from a reference (anchor) location with a single learned scale:

$$k^{EXP}(s) = \exp\left(-\frac{d(s)}{\tau_0}\right), \tag{A.4}$$

where $\tau_0 > 0$ is a learned decay scale and $d(s) \geq 0$ measures distance to the anchor. For $s = \mathbf{x}$, distance is Euclidean, $d(\mathbf{x}_n) = \|\mathbf{x}_n - \mathbf{x}_0\|$, where $\mathbf{x}_0$ is the prediction location and $\mathbf{x}_n$ is a location in the horizontal neighborhood. For $s = p$, distance is defined on the discrete pressure grid. Let $j \in \{0, \ldots, N_p - 1\}$ index pressure levels from top ($j = 0$, 500 hPa) to bottom ($j = N_p - 1$, 1,000 hPa), and define $d(p_j) = (1-\alpha)j + \alpha(N_p - 1 - j)$, where $\alpha \in (0, 1)$ is a learned parameter. This yields a linear interpolation between $j$ (index distance from the top) and $N_p - 1 - j$ (index distance from the bottom), so $\alpha \approx 0$ anchors decay near the top of the column and $\alpha \approx 1$ anchors it near the surface. For $s = t$, distance is defined on the discrete time grid. Let $j \in \{0, \ldots, N_t - 1\}$ index timesteps from the start of the window ($j = 0$ at $t_0 - \tau_{\max}$) to the prediction time ($j = N_t - 1$ at $t_0$), and define $d(t_j) = (N_t - 1) - j$, the number of timesteps before the prediction time, so influence decays exponentially backward into the past.

For numerical stability during optimization, small values ($10^{-8}$) are added to top-hat and mixture-of-Gaussians kernels to prevent exact zeros, and exponential decay scales $\tau_0$ are clamped to the range $[10^{-4}, 100]$. Different parametric kernel families may be assigned to different predictor fields within the same model, allowing the nonlocal aggregation stage to reflect predictor-specific structure while preserving a shared downstream mapping and a common set of kernel families.

## Appendix B: Neural network architecture and training configuration

All models in the hierarchy share an identical downstream neural network that maps either flattened predictor fields (baseline models) or kernel-integrated features (kernel models), together with local inputs, to precipitation. This ensures that differences in predictive performance across models reflect only differences in how nonlocal information is represented, not differences in network architecture or optimization.

The downstream network is a fully connected feedforward model with four hidden layers of widths 256, 128, 64, and 32. Each hidden layer is followed by a GELU activation and dropout with probability 0.1. The output layer consists of a single linear neuron and predicts precipitation in standardized log-transformed space.

Models are trained with a batch size of 500 using the Adam optimizer and an initial learning rate of $5 \times 10^{-4}$. The learning rate is reduced by a factor of 0.5 after two consecutive epochs without validation improvement, with a minimum learning rate of $10^{-6}$. Training proceeds for up to 20 epochs with early stopping based on validation loss, terminating after four consecutive non-improving epochs. The loss function is MSE in this standardized log-transformed precipitation space, and all experiments use fixed random seeds (42, 72, 102) to ensure reproducibility.

## Appendix C: Additional results

***Table C1.*** *Test set performance of models in Figure 2, computed in the standardized log-transformed precipitation space. Results are reported as the mean across seeds, with bootstrap standard deviation (n = 1,000 resamples) and across-seed standard deviation (n = 3) in parentheses.*

| Model | $R^2$ | MSE |
|---|---|---|
| $(\mathbf{x}_0, p_0, t_0)$ | 0.411 ± 0.0015 (0.0006) | 0.680 ± 0.0039 (0.0006) |
| $(\mathbf{x}_0, p_0, t)$ | 0.432 ± 0.0015 (0.0002) | 0.656 ± 0.0037 (0.0003) |
| $(\mathbf{x}, p_0, t_0)$ | 0.440 ± 0.0016 (0.0028) | 0.647 ± 0.0035 (0.0033) |
| $(\mathbf{x}_0, p^{\mathrm{TH}}, t_0)$ | 0.481 ± 0.0015 (0.0014) | 0.599 ± 0.0034 (0.0016) |
| $(\mathbf{x}_0, p^{\mathrm{EXP}}, t_0)$ | 0.482 ± 0.0015 (0.0010) | 0.598 ± 0.0035 (0.0012) |
| $(\mathbf{x}_0, p^{\mathrm{G}}, t_0)$ | 0.484 ± 0.0014 (0.0013) | 0.596 ± 0.0035 (0.0016) |
| $(\mathbf{x}_0, p^{\mathrm{MG}}, t_0)$ | 0.487 ± 0.0015 (0.0003) | 0.592 ± 0.0034 (0.0004) |
| $(\mathbf{x}_0, p^{\mathrm{MIX}}, t_0)$ | 0.488 ± 0.0014 (0.0009) | 0.592 ± 0.0033 (0.0011) |
| $(\mathbf{x}_0, p^{k}, t_0)$ | 0.496 ± 0.0015 (0.0002) | 0.582 ± 0.0033 (0.0003) |
| $(\mathbf{x}_0, p, t_0)$ | 0.528 ± 0.0015 (0.0013) | 0.546 ± 0.0032 (0.0015) |
| $(\mathbf{x}, p, t)$ | 0.582 ± 0.0014 (0.0017) | 0.482 ± 0.0028 (0.0019) |

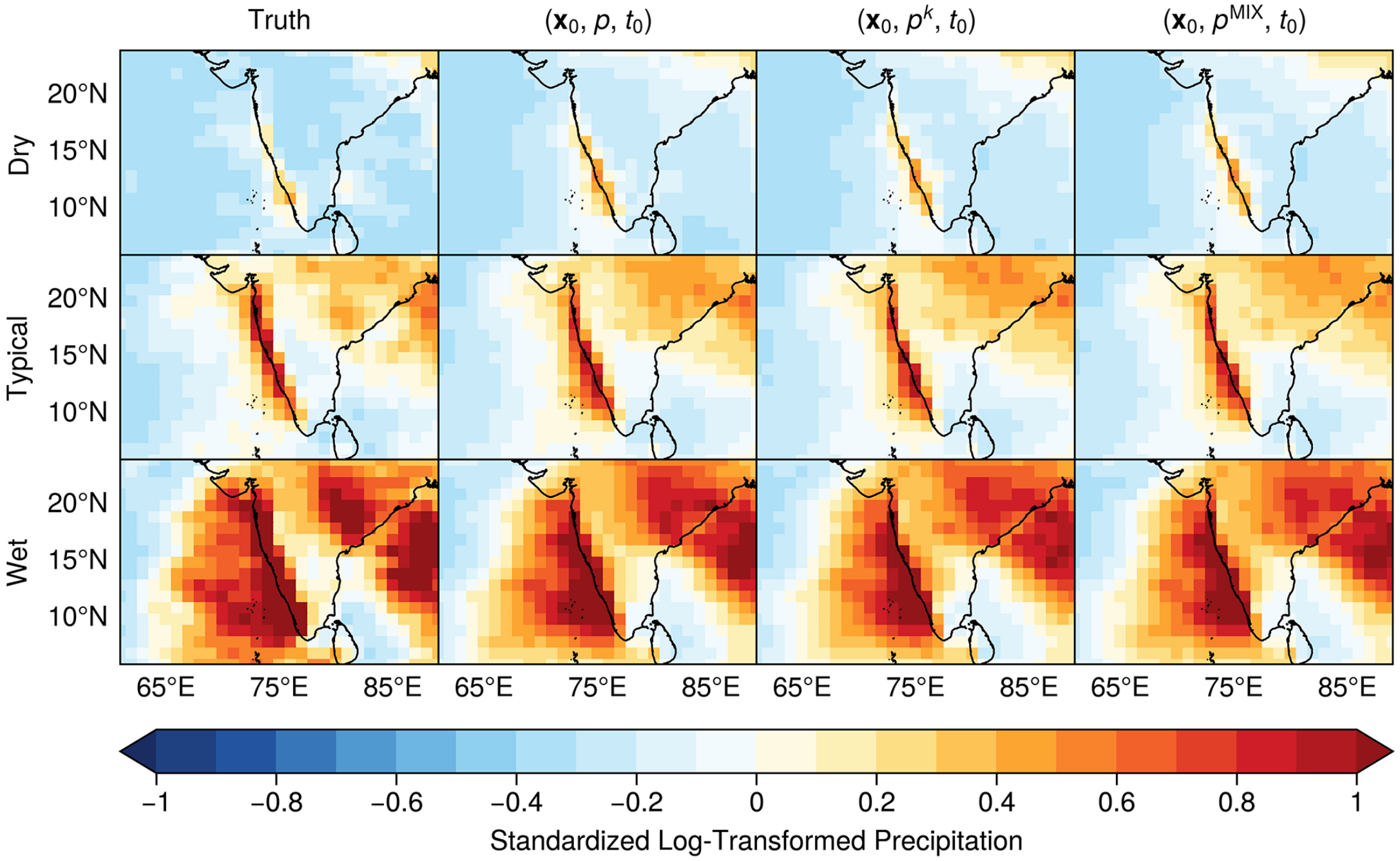

***Figure C1.*** *Spatial composites of standardized log-transformed precipitation for the ground truth and selected baseline and kernel models. Rows show dry (≤ 5$^{th}$), typical (45–55$^{th}$), and wet (≥ 95$^{th}$) percentiles of domain-mean precipitation, averaged over selected timesteps within each regime.*

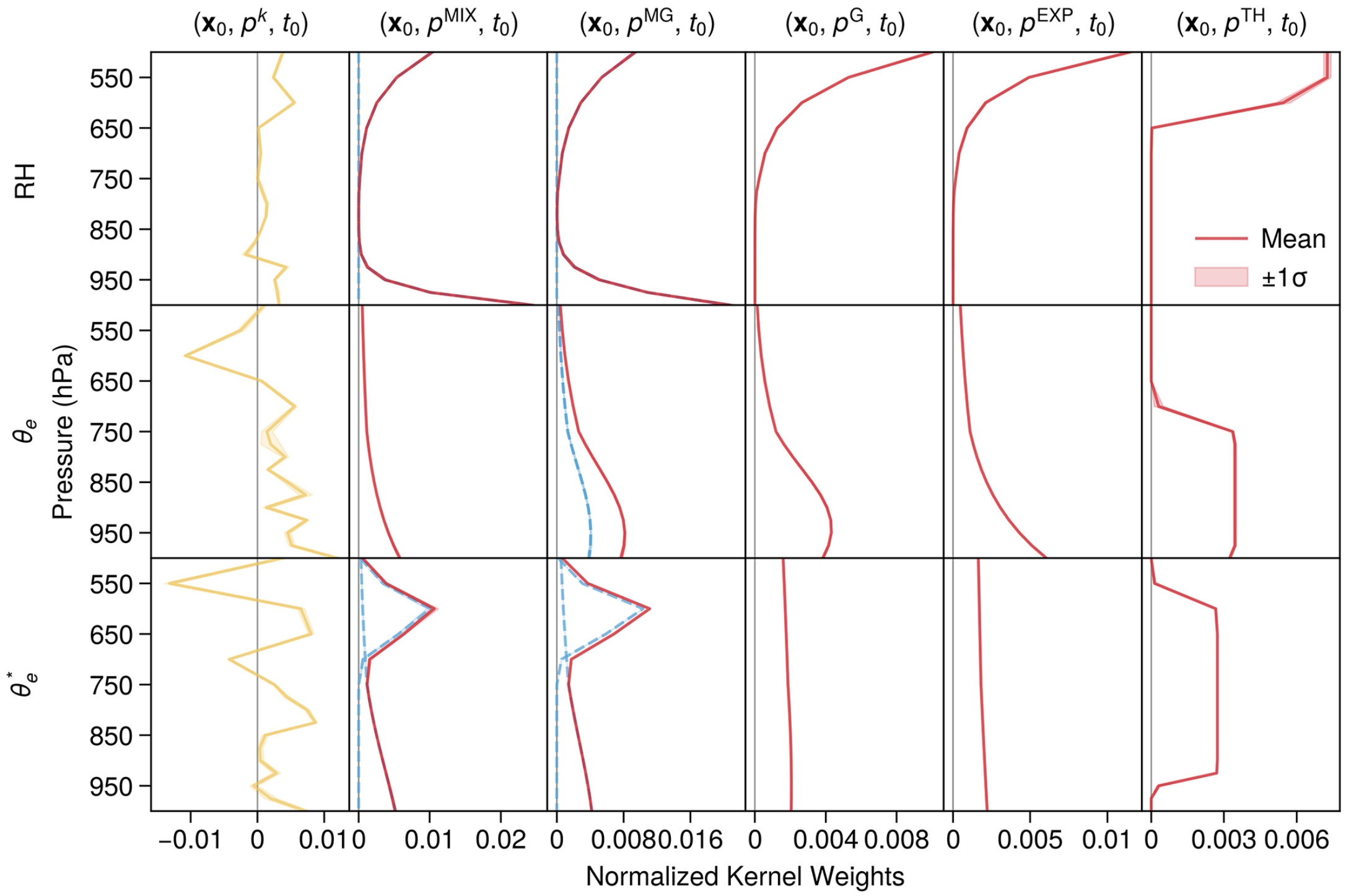

***Figure C2.*** *Learned vertical kernels for all kernel models, extending the results in Figure 3.*